\documentclass[letterpaper, 10 pt, conference]{ieeeconf}  

\IEEEoverridecommandlockouts                              

\usepackage{graphics} 
\usepackage{graphicx}
\usepackage{amsmath} 
\usepackage{amssymb}  
\usepackage{bm}

\usepackage{xspace}
\newcommand*{\eg}{e.g.\@\xspace}
\newcommand*{\ie}{i.e.\@\xspace}

\makeatletter
\let\NAT@parse\undefined
\makeatother

\usepackage{hyperref}

\title{\LARGE \bf
Experimental evaluation of verbal communications in collaborative human-robot transportation
}

\author{Rosanna Coccaro$^{1}$, Enrico Ferrentino$^{1}$, and Pasquale Chiacchio$^{1}$
\thanks{This research was funded by \textit{Istituto Nazionale della Previdenza Sociale} (INPS).}%
\thanks{$^{1}$Authors are with Dept. of Computer Engineering, Electrical Engineering and Applied Mathematics (DIEM),
        University of Salerno, Fisciano, Italy
        {\tt\small rcoccaro@unisa.it}}%
\thanks{©2026 IEEE. Personal use of this material is permitted. Permission
from IEEE must be obtained for all other uses, in any current or future
media, including reprinting/republishing this material for advertising or
promotional purposes, creating new collective works, for resale or
redistribution to servers or lists, or reuse of any copyrighted component
of this work in other works.}
}

\usepackage{eso-pic}
\AddToShipoutPictureBG{%
  \AtTextUpperLeft{%
    \put(0,8){\parbox[b]{\textwidth}{\raggedright\scriptsize
      ACCEPTED VERSION, PUBLISHED AT 10.1109/SMC63593.2026.11788428 - 2026 IEEE INTERNATIONAL CONFERENCE ON SYSTEMS, MAN, AND CYBERNETICS (SMC 2026)}}%
  }%
}

\begin{document}

\maketitle
\thispagestyle{empty}
\pagestyle{empty}

\begin{abstract}

Physical Human-Robot Interaction (pHRI) naturally emerges in collaborative tasks involving force exchange, such as collaborative object transportation, in which robots assist human operators in handling heavy or bulky objects. In such scenarios, force-based information is commonly used to estimate the human's intention of motion. However, relying solely on force-based estimation may lead to ambiguous, if not unpredictable, behaviors, leading to suboptimal coordination and reduced operator comfort. Building upon a baseline force-based interaction framework, the purpose of this paper is thus to experimentally assess to what extent the introduction of an explicit speech-based intention declaration module offers measurable improvements in collaborative transportation. The resulting multimodal human-robot collaboration architecture combines physical interaction forces with spoken commands through an agreement-aware mechanism that safely coordinates verbal and force-based cues. We experimentally evaluate the impact of the additional cognitive channel on a collaborative transportation task involving an industrial robot and multiple human subjects, comparing the baseline force-based information only system against the speech-augmented version. The evaluation combines both objective performance metrics and subjective user feedback collected via questionnaires. The results demonstrate that the introduction of explicit intention declaration through speech reduces user effort, while preserving overall comfort during task execution. These findings prove that adding a cognitive communication channel can effectively complement physical interaction and improve performance and interaction quality in collaborative transportation, leading to a more intuitive and comfortable human-robot collaboration.

\end{abstract}

\section{INTRODUCTION}
\label{sec:introduction}

Physical Human-Robot Collaboration (HRC) has gained significant attention within contemporary industrial contexts \cite{ajoudaniProgressProspectsHuman2018}, where robots and humans share the same workspace, with the former supporting human operators in tasks involving manipulation and transportation in dynamic and semi-structured environments \cite{gualtieriOpportunitiesChallengesSME2020, hentoutHumanRobotInteraction2019}.

In many collaborative transportation applications \cite{hamad_2021, park_2024}, human intentions are typically inferred from physical action forces applied to the robot directly or through a shared object. Accordingly, interaction (admittance, impedance) control has been the predominant approach to manage physical Human-Robot Interaction (pHRI) \cite{ajoudaniProgressProspectsHuman2018}. However, a limitation of existing force-based interaction strategies lies in their reactive nature, as the robot initiates the motion only once the user applies sufficient physical force. Moreover, direct force feedback is not always available, such as when the manipulated object is deformable \cite{Sirintuna_2023} or when tasks require frequent direction changes or anticipatory movements. Under these conditions, task execution may be significantly compromised, potentially preventing successful completion altogether. Therefore, it is beneficial if the robot could receive contextual awareness through complementary modalities \cite{solak_2025, hayashida_2022, lukac_2017, olivas_2023}.

Multimodal HRC approaches have explored integrating additional communication channels, including hand gestures, eye gaze \cite{wang_2025, xu_2023}, brain-computer interfaces \cite{zhang_2022}, haptic-visual fusion \cite{agravante_2014} and voice commands \cite{rosen_2020} to complement force-based interaction. Among these, language-based modalities, and large language models (LLMs), in particular, have emerged as a promising direction for HRC, offering natural language understanding, high-level reasoning, and plan generation capabilities \cite{zhang_2023}. Speech, indeed, stands out as a natural and expressive modality that allows humans to explicitly declare their intended actions. Moreover, it is particularly well suited for manipulation tasks, as it allows users to keep their hands free for interaction with objects. However, despite the growing interest in multimodal HRC, quantitative assessments of how explicit speech-based intention declaration affects physical user effort and comfort in assistive transportation tasks remain limited, as current evaluations primarily focus on interface usability and task precision rather than on the reduction of physical energy expenditure. While the studies that do measure objective physical effort (\eg, electromyography or metabolic cost) are still a few, they either analyze a single isolated aspect of effort or rely on short-term tasks confined to controlled laboratory environments \cite{su_2023, Strazdas_2025, kalatzis_2023, zakeri_2023}.

With this paper, we provide an objective characterization of performance in a pHRI task when a cognitive communication channel is introduced. We propose a multimodal HRC architecture that extends our baseline Assistive Force Control (AFC) framework \cite{2023_lima} by integrating speech as an additional communication modality. The system combines physical interaction forces and spoken commands through an agreement-aware mechanism that evaluates the consistency between verbal and physical cues, ensuring that the robot delivers a safe and reliable assistive behavior. The architecture introduces several key functionalities, including pre-engagement (PE) assistance activated by spoken intentions and disagreement detection to safely stop the interaction when inconsistencies are detected. We note that the primary objective of this work is to investigate the integration of speech as a cognitive communication channel within the proposed pHRI and assistive control framework. Therefore, the optimization of the speech recognition system, e.g., in terms of robustness to noise, less rigid vocabulary, accented speech, lies beyond the scope of this paper.

The proposed approach is then validated through experiments in which humans perform a collaborative transportation task with an industrial robot. Two interaction modalities are compared: Force-Based only (FB) and Force-Based with Voice (FBV). The evaluation includes objective performance metrics quantifying user physical effort (as the work index $i_W$ and the force index $i_F$), as well as subjective measures of comfort and perceived workload from post-experiment questionnaires.

\section{METHODOLOGY} \label{sec:methodology}

The baseline system provides assistance in terms of assistive force references during physical human-robot collaboration by estimating the human partner's intention of motion from interaction forces. Forces measured at the end-effector are preprocessed to remove bias, reduce noise, and filter small disturbances, following approaches such as \cite{cherubini_2016}, where smoothing filters and dead-band techniques were employed. Building upon this framework, we extend such a baseline by integrating a speech-based cognitive communication channel and a multimodal intention fusion strategy, as described in the following sections.

\subsection{Multimodal architecture overview}

\begin{figure*}
  \centering
   \includegraphics[width=\textwidth]{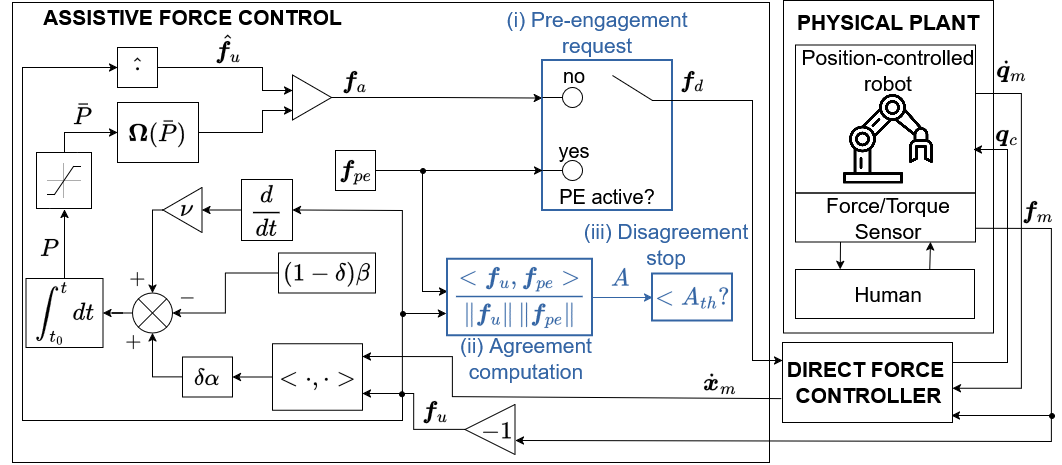}
  \caption{Block diagram of the proposed architecture showing, in blue, the differences with respect to the baseline architecture, \ie, (i) pre-engagement assistance (Section \ref{sec:pre_engagement_assistance}), (ii) agreement computation (Section \ref{sec:agreement_aware_control}), and (iii) disagreement stop (Section \ref{sec:agreement_aware_control}).}
  \label{fig:assistive_force_control}
 \end{figure*}

The overall architecture of the proposed system, inspired by \cite{2023_lima}, is shown in Figure \ref{fig:assistive_force_control}. The framework is composed of an AFC module and a Direct Force Controller (DFC). The AFC operates estimating the human partner's intention of motion and generating the desired assistive force references. The DFC, implemented as a low-level controller for a position-controlled manipulator, converts these force references into smooth position commands, thereby enabling compliant and robust physical human-robot interaction. 

Therefore, within the AFC, interaction forces measured at the end-effector are continuously processed and used to estimate the user's intended direction of motion. The assistive force is computed along this direction and modulated in magnitude through a passivity-based mechanism.

In parallel, a speech command recognition module allows the user to explicitly declare a desired direction of motion using specific commands. Speech input enables anticipatory assistance and complements force-based intention estimation, particularly in situations where interaction forces are weak or delayed. The fusion of physical and verbal cues is handled by an agreement-aware control mechanism, which evaluates the consistency between the direction inferred from measured forces and the direction declared through speech. When sufficient agreement is detected, within a predefined time window, the robot provides assistive forces in the intended direction of motion. Conversely, in the presence of a disagreement, assistance is reduced or halted to preserve safety and interaction transparency.

\subsection{Force-based intention estimation and assistance}

The user's intention of motion is quantified through a passivity index $P(t)$, which modulates the magnitude of the robot's assistive force, rising when the user applies forces and gradually decaying when they cease, enabling smooth and intuitive assistance.

Building on the original formulation in \cite{ikeura_cooperative_1995} and its extension in \cite{2023_lima}, we adopt:
\begin{equation}
    \begin{aligned}
    P(t) = \int_{t_0}^t  \Big( 
    & \delta(t) \alpha \langle \dot{\bm{x}}_m(t), \bm{f}_u(t) \rangle \\
    & - \big(1 - \delta(t)\big) \beta - \nu \lVert \dot{\bm{f}}_{u}(t) \rVert \Big) dt,
    \end{aligned}
\end{equation}
where $\alpha, \beta \in \mathbb{R}^{+}$ are positive scalars regulating the rise and decay of assistance, respectively, $\dot{\bm{x}}_m(t) \in \mathbb{R}^3$ is the end-effector velocity, $\bm{f}_m(t) \in \mathbb{R}^3$ is the robot-applied force by its end-effector over the object/environment, $\bm{f}_u(t) = - \bm{f}_m(t)$ is the user-applied force, $\delta(t) = 1$ if $\lVert \bm{f}_m \rVert > 0$, and $\delta(t) = 0$, otherwise.
The damping constant $\nu$ is introduced to enable energy dissipation when forces are exchanged between the human and the robot \cite{hogan_84}.

The passivity index is clamped to $[0,1]$ and mapped to the assistive force magnitude via the linear activation function $\Omega_\gamma(\bar{P}) = \gamma \bar{P}(t): [0, 1] \to [0, \gamma]$, where $\gamma$ is the maximum allowable assistive force norm and $\bar{P}$ is the clamped passivity index. The resulting assistive force is therefore defined as:
\begin{equation} \label{eq:assistive_force}
    \bm{f}_a(t) = \bm{\hat{f}}_u(t) \Omega \big( \bar{P} (t) \big),
\end{equation}
where $\bm{\hat{f}}_u(t)$ is the direction of the user-applied force, and the DFC reference is $\bm{f}_d = \bm{f}_a$. This formulation ensures that assistance is provided in the direction intended by the user while maintaining smooth and stable interaction behavior.

\subsection{Cognitive human-robot interaction}

\subsubsection{Speech command recognition}

The speech command recognition system operates through a four-stage pipeline: audio acquisition, voice activity detection, speech transcription, and command matching. First, audio signals are captured in real-time from a microphone connected to the system. Then, detected speech segments are transcribed to text using a pre-trained Wav2Vec2 neural network with Connectionist Temporal Classification (CTC) decoding \cite{K_rzinger_2020}. The system utilizes a pre-trained model from the SpeechBrain framework \cite{speechbrain_v1}, specifically trained on the Common Voice English dataset \cite{ardila_2020_common}. The audio is first normalized, then processed through the neural network to generate the text transcription. Finally, the transcribed text is matched against a predefined set of valid commands. The matching strategy used in this work adopts the Character Error Rate (CER) as the similarity metric. The system selects the best-matching command based on error scores, returning an ``unknown'' classification if no command meets the minimum confidence threshold.

CER is defined as:
\begin{equation}
    CER = \frac{S + D + I}{M}
\end{equation}
where $S$, $D$, and $I$ denote the number of character-level substitutions, deletions, and insertions, respectively, and $M$ is the total number of characters in the reference string. CER is computed via the torchmetrics library \cite{Detlefsen2022} and yields a value in $[0,1]$, where 0 indicates a perfect match. For each candidate command, the CER between the recognized input text and the command string is computed, and the command with the lowest CER, provided it remains below the configured threshold $CER_{max}$, is selected as the recognized command.

This approach requires no machine learning model or GPU, making it lightweight and suitable for deployment in resource-constrained or real-time robotic systems. The output of the speech module is a discrete motion intention label, which is mapped to a unit direction vector representing the expected user-applied force in the user's reference frame.

\subsubsection{Text-to-speech feedback system}

Additionally, the system provides vocal feedback to the user through a text-to-speech (TTS) module that converts textual responses into highly natural and expressive synthesized speech. The implementation is based on the Coqui TTS engine \cite{Eren_Coqui_TTS_2021} within a client–server architecture for speech synthesis. The server is deployed in a Docker container and hosts a Variational Inference Text-to-Speech (VITS) \cite{kim_2021} model. When auditory feedback is required, the desired text is sent to the server via an HTTP request. The server generates the corresponding speech signal and returns it as a .WAV file, which is then played by the client. This design decouples the computationally intensive synthesis process from the recognition pipeline, enabling GPU-accelerated TTS generation while preserving real-time responsiveness.

\subsection{Multimodal Intention Fusion}

Speech-based intention declaration offers two key benefits. First, it supports pre-engagement assistance, where the robot begins moving in the declared direction before any physical interaction occurs. Second, it provides an explicit reference direction that can be compared with measured interaction forces, allowing detection of agreement or disagreement between verbal and physical cues.

Importantly, speech input does not override physical interaction. Rather, it acts as a complementary source of intention information, which is integrated with force measurements through the agreement-aware control strategy described next.

\subsubsection{Pre-engagement assistance}  \label{sec:pre_engagement_assistance}

In human-human collaborative transportation, explicitly declaring motion intent facilitates coordination. We propose to make the cognitive information flow on the physical assistance during human-robot collaboration in the form of anticipatory forces.

Therefore, we introduce the concept of explicit pre-engagement (PE) assistance, where the robot initiates movement in anticipation (or at the same time) of the user's force application. This replaces the standard computation of the DFC desired force $\bm{f}_d$ upon receiving a speech command (\emph{PE active} condition). Here, the robot generates limited assistive force in the declared direction $\hat{\bm{f}}_{pe}$ with magnitude $\lVert \bm{f}_{pe} \rVert$, before or meanwhile the user physically applies forces. Once $\lVert \bm{f}_u \rVert \ge \lVert \bm{f}_{pe} \rVert$, control reverts to standard force-based assistance, i.e., 
\begin{equation}
\bm{f}_d(t) =
    \begin{cases}
    \bm{f}_{pe}, & \text{if \emph{PE active} AND}\, \lVert \bm{f}_u(t) \rVert < \lVert \bm{f}_{pe} \rVert,\\
    \bm{f}_{a}(t), & \text{otherwise}. \\
    \end{cases}    
\end{equation}
We note that the switching criterion only considers the force magnitude $\lVert \bm{f}_u \rVert$, not its direction. Thus, sufficient force in any direction deactivates pre-engagement and returns to standard force-based assistive control, i.e., physical forces preempt the spoken intention. This ensures flexibility to contingencies and improves system safety, while preventing the user from perceiving discomfortable discontinuities in the assistive force.

\subsubsection{Agreement aware control} \label{sec:agreement_aware_control}

To coherently integrate speech-based intention declaration with force-based intention estimation, we introduce an agreement-aware control mechanism that regulates the assistive behavior of the robot based on the consistency between verbal and physical cues.

When a speech command is issued, it generates an expected force direction $\hat{\bm{f}}_{pe}$. During physical interaction, the user-applied force $\bm{f}_u$ is continuously monitored, and the agreement between the declared and measured directions is quantified using cosine similarity:
\begin{equation} \label{eq:agreement}
    A = \frac{\left< \bm{f}_u, \bm{f}_{pe} \right>}{\lVert\bm{f}_u \rVert \lVert \bm{f}_{pe} \rVert},
\end{equation}
where $A \in [-1, 1]$. $A$ close to 1 indicates strong alignment, whereas values near $0$ and negative respectively indicate weak agreement or disagreement (opposing directions).

When $A$ exceeds a predefined agreement threshold $A_{th}$, the verbal and physical cues are considered consistent. In this case, assistive forces are generated in the declared direction of motion, combining the benefits of anticipatory speech input with continuous force-based modulation through the passivity index $P(t)$. Conversely, if $A<A_{th}$, the system detects an inconsistency between declared and measured intentions. Such disagreement may arise due to incorrect spoken intention or change of user intention. In this situation, the assistive force is halted to prevent unintended motion in a direction not physically supported by the user. This strategy preserves safety, guaranteeing that assistance remains coherent with the user's actual behavior.

Since PE assistance plays the role to provide anticipatory assistive forces in the intended spoken direction of motion, with the ultimate goal to facilitate motion initiation in the handling of bulky or heavy objects, agreement monitoring is deactivated after a configurable time threshold $t_{th}$. After this time has elapsed, the user can freely change direction, without incurring in the disagreement condition.

\subsection{Quantitative metrics for performance characterization}

To quantitatively assess system performance, we define the following metrics.

\subsubsection{Work index}

The total absolute work performed by the user can be approximated with
\begin{equation}
    i_W = \sum_{i=1}^N \left| \left< \bm{f}_u(i), \Delta \bm{x}_m (i) \right> \right|,
\end{equation}
where $N$ is the total number of discrete timestamps of the recorded task and $\Delta \bm{x}_m(i)$ is the position displacement. The work index approximates the user's metabolic energy expenditure and fatigue, as measured on the robot side, if user-applied forces are fully transmitted through the rigid object being manipulated. The higher the $i_W$, the greater the user effort during interaction.

\subsubsection{Force index}

The force index is defined as
\begin{equation}
    i_F = \frac{1}{\Lambda} \sqrt{\frac{\sum_{i=1}^N \lVert \bm{f}_u(i)\rVert^2}{N}},
\end{equation}
where $\Lambda$ is the total transportation path length. The index \texorpdfstring{$i_{F}$}{i\_F} is normalized by trajectory length and is independent of its shape. Higher \(i_F\) values indicate reduced robot assistance efficiency, requiring greater user force to achieve motion.

\section{EXPERIMENTS}

\begin{figure}
  \centering
   \includegraphics[width=.47\textwidth]{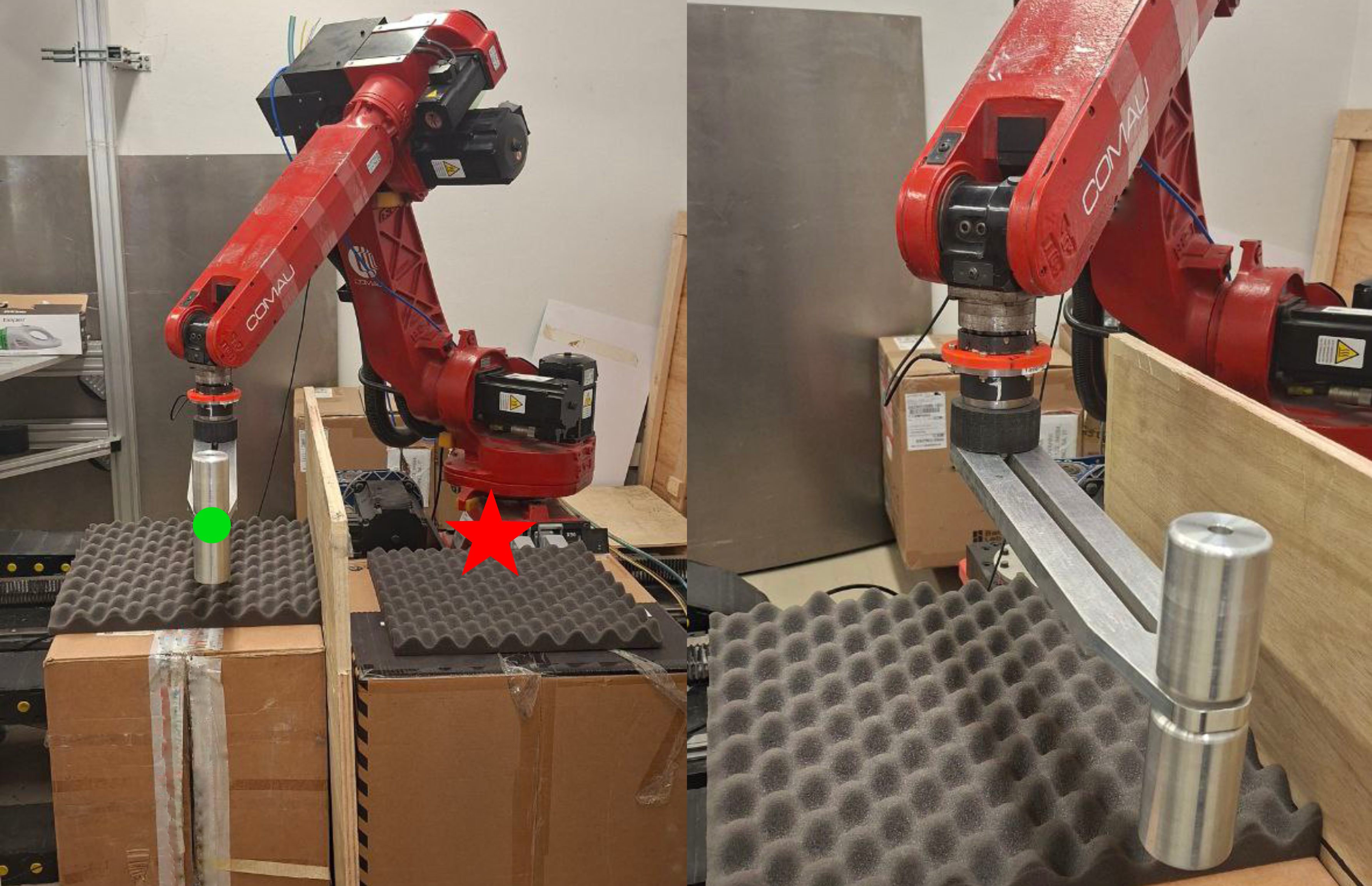}
  \caption{Experiment setup, showing the two target positions involved in the experiment on the left (green dot is the left target, red star is the right target) and a more detailed view of the metal plate on the right. The objective is to transport the metal plate from the left to the right target. Then, the participant is asked to make the path in the reversed way.}
  \label{fig:experiments}
 \end{figure}

The objective of the experimental evaluation is to assess the impact of speech-based intention declaration on user effort, motion quality, and perceived comfort during collaborative transportation tasks. The evaluation is designed to test two hypotheses: first, that the integration of speech reduces the physical effort required from the user (as reflected by lower force and work indices), and second, that speech-based intention declaration preserves subjective safety, predictability, naturalness, and comfort.

For such validation, participants are recruited among the students and employees of the University of Salerno. All subjects provide informed consent by reading and signing a document outlining the study's purpose and the participants' rights\footnote{The experimental procedure had no potential to damage the mental or physical health of human participants. The research project did not involve vulnerable participants, nor did it have access to private or sensitive data for purposes that are not related to the project. The authors adhered to the code of conduct of the University of Salerno.}. In the end, 23 subjects, aged between 20 and 42 (\textit{mean} $26.75$, \textit{std} $\pm 5.03$), are enrolled.

We employ the position-controlled industrial robot Comau SmartSix 6-1.4, which supports payloads up to 6 kg, and a Schunk FTD MINI58-IP60 SI-700-30 force/torque sensor. The robot transported object (the metal plate, mass $\sim2$ kg, size $43 \times 7 \times 8$ cm) is held by the robot through a mechanical fuse, designed to break in case of uncontrolled collisions.

The task consists of collaboratively transporting the metal plate across the robot workspace, repeatedly moving it from one side to the other and back, while avoiding a vertical obstacle positioned in the middle. The task requires coordinated motion in the vertical plane, with constant end-effector orientation throughout the task.

We set the parameters of the system as in Table \ref{tab:controller_parameters}, tuned empirically, through preliminary tests. The same physical setup is employed for both FB and FBV experimental modalities to ensure comparability. Each of the 23 participants is requested to carry out the task in both experimental modalities. For each modality, the experiment is designed to last 5 minutes, during which participants repeatedly perform the transportation task between the two target positions. We limit experiments to 5 minutes so to prevent tiredness factors to arise and consequently affect the user's comparisons between modalities.

\begin{table}[h!]
\centering
\caption{Assistive Force Controller Parameters Utilized in the Experimental Setup}
\label{tab:controller_parameters}
\begin{tabular}{ccll}
\hline
\textbf{Parameter} & \textbf{Value} & \textbf{Description} \\ \hline
$\alpha$ & $1$ & Passivity assistance rise factor \\
$\beta$ & $1$ & Passivity assistance decay factor \\
$\nu$ & $0.07$ & Energy dissipation damping constant \\
$P_{th}$ & $0.5$ & Passivity index activation threshold \\
$\gamma$ & $150$ N & Maximum assistive force \\
$\lVert \bm{f}_{pe} \rVert$ & $15$ N & Pre-engagement assistive force \\
$A_{th}$ & $-0.9$ & Agreement threshold for disagreement detection \\
$t_{th}$ & $2$ s & Time threshold for disagreement detection \\
$CER_{max}$ & $0.6$ & CER Threshold \\
\hline
\end{tabular}
\end{table}

Prior to each modality, participants receive a standardized verbal explanation of the procedure and observe a live demonstration of the system. They are also given the opportunity to freely test the setup to ensure a clear understanding of the task requirements and to become familiar and confident with the system. The main evaluation then follows a fixed order, with FB always performed before FBV, and each block limited to 5 minutes to minimize fatigue. For the FBV modality, participants are additionally provided with a written protocol (see Appendix \ref{app:protocol}) and a cheat sheet listing the available voice commands. These commands allow them to specify the intended direction of motion, activate or deactivate voice recognition, and start or stop the force controller. This ensures consistent instruction across participants while allowing natural interaction during the task. After completing each modality, participants fill out a questionnaire assessing perceived safety, predictability, naturalness, and comfort. Speech-specific questions appear only in the FBV condition.

The list of questions for each modality is based on an objective questionnaire. Answers are modeled according to a 5-points Likert scale, including redundant questions regarding human general feeling, assistive force system, standalone speech recognition, and speech recognition applied to AFC. We use redundant questions to identify consistent answers to be included in our statistical analysis. The administered questionnaire is available in Appendix \ref{app:questionnaire}.

\section{RESULTS AND DISCUSSION}

\begin{table}
\caption{Mean and standard deviation of $\lVert \bm{f}_u \rVert$, $i_W$, $i_F$ and task duration for both FB and FBV modalities}
\label{tab:metrics}
\begin{center}
\setlength{\tabcolsep}{4pt}
\begin{tabular}{l|c|c}
Metric              & FB                            & FBV\\ \hline
User Force ($\lVert \bm{f}_u \rVert$)     & $23.23 \pm 8.71$ N            & $5.21 \pm 2.50$ N\\
Work Index ($i_W$)  & $535.84 \pm 327.93$ N$\cdot$m & $96.69 \pm 68.01$ N$\cdot$m\\
Force Index ($i_F$) & $1.41 \pm 0.41$ N             & $0.67 \pm 0.28$ N \\
Task Duration       & $183.14 \pm 84.10$ s          & $231.29 \pm 88.38$ s
\end{tabular}
\end{center}
\end{table}

Table \ref{tab:metrics} provides a comparative analysis of FB and FBV modalities through the mean and standard deviation of user force ($\lVert \bm{f}_u \rVert$), work Index ($i_W$) and force Index ($i_F$). Let $\Delta$ be the average difference between the FB and FBV conditions (FB $-$ FBV), while Cohen's $d$ quantifies the standardized effect size of each difference. These values indicate that speech integration substantially reduces the mechanical load imposed on the human partner, with mean user force decreasing by $77.6\%$ ($\Delta = 18.02$~N, 95\% CI $[15.25, 20.79]$, $d = 2.81$), $i_W$ by $82.0\%$ ($\Delta = 439.15$~N$\cdot$m, 95\% CI $[336.7, 541.6]$, $d = 1.85$), and $i_F$ by $52.5\%$ ($\Delta = 0.74$~N, 95\% CI $[0.59, 0.90]$, $d = 2.12$). All three differences were confirmed significant by paired $t$-tests and Wilcoxon signed-rank tests ($p < .0001$ for the entire dataset), with very large effect sizes ($d > 1.8$ for all metrics), indicating that these reductions are not only statistically reliable but also practically substantial. From an ergonomic standpoint, such reductions are highly relevant for repetitive or prolonged collaborative tasks, where cumulative physical effort can lead to fatigue and musculoskeletal strain. These findings are further supported by users reporting that ``less force is required'' and the ``absence of fatigue'' in the FBV condition, although user's fatigue is expected to increase over time. An inversion in the order of experimental conditions would further exacerbate this difference. Speech integration also improves interaction consistency, as the passivity index decreased by $61.2\%$, suggesting that explicit intention declaration reduces ambiguity in motion initiation and direction selection. This, in turn, directly contributes to the observed decrease in physical effort. 

Overall, based on subjective questionnaires, participants rated both conditions positively across all dimensions: they reported feeling safe ($4.41 \pm 0.73$ FBV, $4.61 \pm 0.50$ FB), finding the robot's behavior predictable ($4.45 \pm 0.86$ and $4.38 \pm 0.59$, respectively), and perceiving its motion as natural ($3.96 \pm 0.81$ and $3.96 \pm 0.69$, respectively) and comfortable ($4.19 \pm 0.93$ and $4.50 \pm 0.66$, respectively). Wilcoxon signed-rank tests showed no significant differences between conditions on any item (all $p > 0.05$), indicating speech integration preserved perceived safety, predictability, naturalness, and comfort. As far as the perceived effort is concerned, on a par with the other indices, there is no statistical significance in our population between the FB and FBV conditions. The inconsistency between subjective and objective indices may confirm that the voice integration might become subjectively convenient only after prolonged interactions. Within the speech condition, participants further stated that the vocabulary was easy to remember (88\%) and direction intent easy to express (84\%). On the other hand, the FBV modality resulted in $30.9\%$ increase in task duration. 


Overall, out of the 23 participants, 60\% (14 users) preferred the interaction with voice recognition, but ultimately, the validation presents two different styles of collaboration: the FB modality prioritizes speed and pure physical control at the cost of higher user effort, whereas the FBV modality prioritizes reduced effort and intent-alignment at the cost of temporal efficiency. This 30.9\% increase of task duration in the FBV mode is primarily driven by the inherent latency of the communication pipeline, accounting for the time required for the user to speak and the subsequent assistive force controller actions. Furthermore, the disagreement stop, while improving safety, introduces interruptions whenever voice commands and physical force measurements diverge. For applications involving prolonged repetitive tasks or heavy objects, however, the 77.6\% reduction in user force may well justify this temporal overhead.

This overhead is best interpreted in the context of flexible industrial settings, where heavy, bulky, or irregular loads require collaborative intervention of two operators, or more. In such scenarios, replacing a two-operator manual transport task with a single human-robot pair can improve labor allocation, as the second operator can be reassigned to other tasks. More importantly, the 77.6\% reduction in applied force and 82.0\% reduction in work index suggest a substantial decrease in cumulative musculoskeletal strain, which is particularly valuable in repetitive or prolonged operations. For these two reasons, in non-structured collaborative transportation, the ergonomic and organizational benefits of voice-coordinated assistance can outweigh the increase in execution time of a single operation.

\begin{figure}
  \centering
   \includegraphics[width=.47\textwidth]{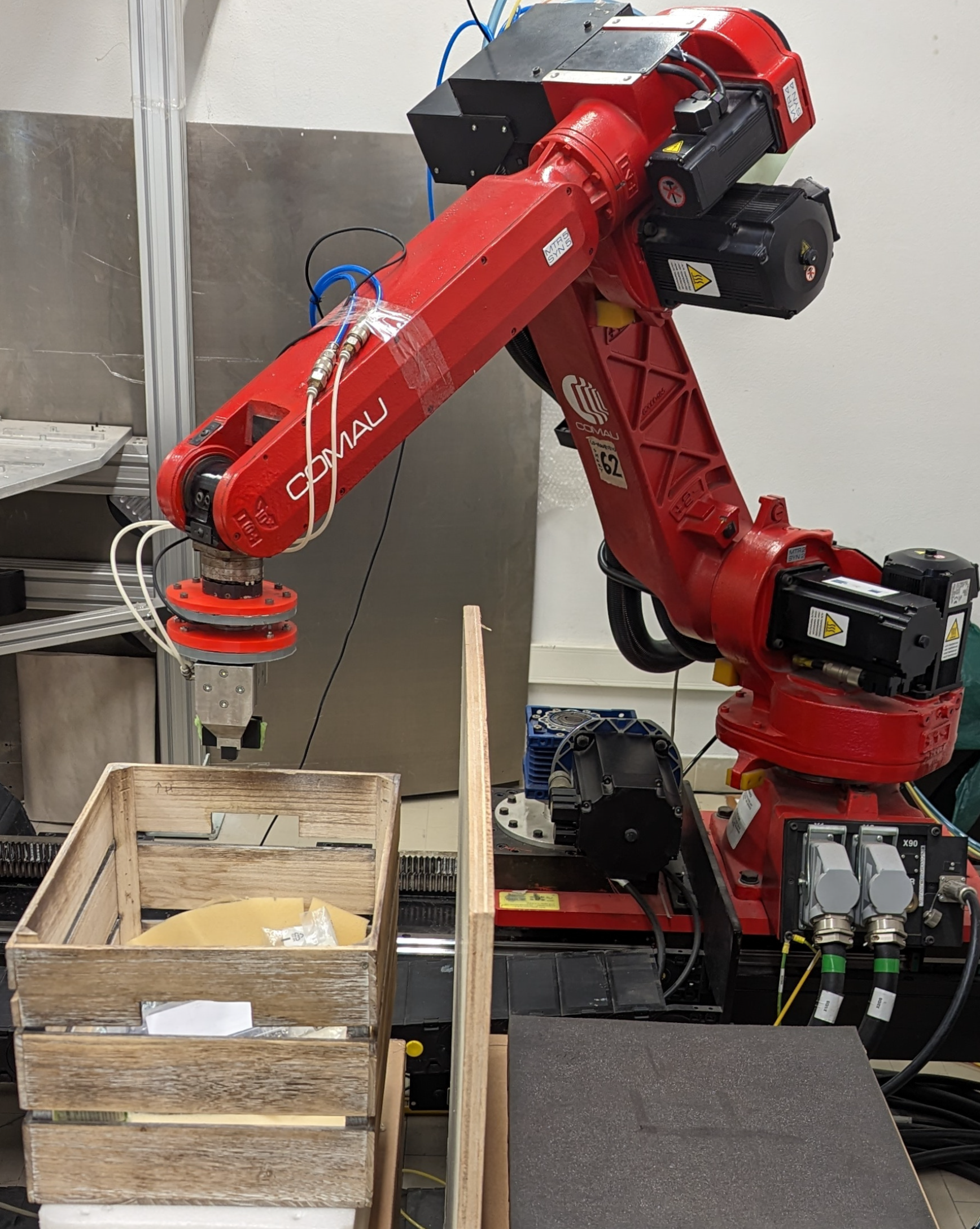}
  \caption{Wooden box experiment setup.}
  \label{fig:wooden_box}
 \end{figure}

To carry out a preliminary assessment of the framework's behavior in the edge case of manipulating a medium-flexible object, we conducted additional tests with 5 participants in which the metal plate was replaced with a wooden box grasped by a two-finger pneumatic gripper (see Figure \ref{fig:wooden_box}). While the gripper itself provides a rigid grasp once closed (no slippage of the box), the two-finger contact allows rotations about the contact point (roll, pitch, and yaw), and the box's own structural compliance introduces a further source of flexibility not present in the main metal plate setup. The small number of trials conducted in this setup does not allow us to carry out formal statistical analyses. Nonetheless, none of the participants could complete the task as efficiently without the speech channel, indicating that explicit intention declaration may become crucial for ergonomic task completion when force transmission is attenuated by grasp and/or object compliance.

\section{CONCLUSIONS AND FUTURE WORKS}
\label{sec:conclusion}

This work investigated the integration of speech-based intention declaration within an agreement-aware assistive force controller for collaborative object transportation.

A controlled user study with 23 participants demonstrated that explicit verbal intention significantly reduces physical effort during human-robot collaboration. The FBV modality produced substantial reductions in mean applied force ($77.6\%$) and work ($82.0\%$), accompanied by reduced passivity. These results were not supported by subjective assessments, possibly suggesting that benefits become perceivable only after prolonged interactions. Nonetheless, the integration of a cognitive voice channel did not negatively affect safety, predictability, naturalness and comfort, as perceived by our population of users, with the 60\% of them expressing an explicit preference towards the FBV modality.

Although speech integration resulted in slower task execution, users perceived the interaction speed as adequate ($4.21 \pm 0.59$), suggesting that predictability and reduced physical strain may be more critical than raw execution speed in collaborative transportation tasks.

Overall, the findings indicate that explicit multimodal agreement mechanisms do not just improve intention estimation accuracy, but change interaction dynamics toward more sustainable and effort efficient collaboration. This is particularly relevant for assistive, rehabilitation, and repetitive industrial scenarios where long-term physical burden is a critical factor.

Future work will extend the analysis of more complex and realistic task configurations, such as the manipulation of non-rigid or loosely held objects, only preliminarily addressed in this work. Additionally, the current study did not assess how participant characteristics such as age, domain expertise, or prior familiarity with robotic and speech-based systems affect acceptance of the proposed framework and tolerance for the execution delays introduced by the speech channel. Future work will involve participants with more diverse age ranges and industrial expertise, to better characterize these effects under realistic workforce conditions. Finally, more advanced speech recognition models will be assessed, enabling users to communicate their intentions through natural language instead of predefined commands, expected to further reduce the required cognitive effort.


\appendices

    \section{Protocol - Experiment of collaborative transportation with speech recognition}
    \label{app:protocol}
    
    \noindent\emph{To start, you can say:}
    \smallskip
    
    {\centering \textbf{\textit{HEY ROBOT}}\par}
    
    \smallskip
    \noindent\emph{To have the robot follow and assist you, you can say:}
    \smallskip
    
    {\centering \textbf{\textit{FOLLOW ME}}\par}
    
    \smallskip
    \noindent\emph{Wait for confirmation that the command has been accepted, then you may start moving.}
    
    \smallskip
    \noindent\emph{During your movements (up, down, left, and right), you may explicitly declare your motion intention, if you deem it appropriate:}
    \smallskip
    
    {\centering
    \textbf{\textit{MOVE UP / MOVE DOWN}}\\
    \textbf{\textit{MOVE LEFT / MOVE RIGHT}}\par}
    
    \smallskip
    \noindent\emph{WARNING: if the robot detects a mismatch between the declared motion intention and the actual motion, it will stop as a precautionary measure.}
    
    \smallskip
    \noindent\emph{Once you reach a target position of your choice, you may wait briefly and then resume. You can repeat this movement from one target to another and back several times; there is no limit. The total duration of the experiment is 5 minutes.}
    
    \smallskip
    \noindent\emph{If, for any reason, you wish to stop the robot, you can say:}
    \smallskip
    
    {\centering \textbf{\textit{STOP}}\par}
    
    \smallskip
    \noindent\emph{To interrupt voice command recognition at any point during the experiment, you can say:}
    \smallskip
    
    {\centering \textbf{\textit{JUST A MOMENT}}\par}

\section{Post-Interaction Questionnaire}
\label{app:questionnaire}

\small
\smallskip
\noindent\textbf{Safety}
\begin{enumerate}
    \item I felt safe when working with the robot
    \item I was afraid to get hurt or to cause damages
    \item I felt the same level of safety from the start until the end
    \item I felt more safe the more I repeated the experiments
\end{enumerate}

\noindent\textbf{Predictability}
\begin{enumerate}
    \setcounter{enumi}{4}
    \item The robot's behavior was consistent
    \item The robot behaved unexpectedly
\end{enumerate}

\noindent\textbf{Comfort}
\begin{enumerate}
    \setcounter{enumi}{6}
    \item I felt comfortable using the system as a whole
    \item The system complicates more than it helps
    \item I felt more comfortable the more I repeated the experiments
    \item I felt the same level of comfort from the start until the end
\end{enumerate}

\noindent\textbf{Naturalness}
\begin{enumerate}
    \setcounter{enumi}{10}
    \item The robot's motion felt natural
    \item The starting motion of the robot was unnatural
    \item The stopping motion of the robot was unnatural
\end{enumerate}

\noindent\textbf{Velocity}
\begin{enumerate}
    \setcounter{enumi}{13}
    \item The robot's velocity felt just right
    \item The robot was too fast
    \item The robot was too slow
\end{enumerate}

\noindent\textbf{Physical effort}
\begin{enumerate}
    \setcounter{enumi}{16}
    \item I could move the object without feeling tired
    \item I had to apply much force to move the object
    \item I was satisfied with the assistance provided by the robot when I applied little force
    \item I was satisfied with the assistance provided by the robot when I applied a lot of force
\end{enumerate}

\noindent\textbf{Communication (FBV only)}
\begin{enumerate}
    \setcounter{enumi}{20}
    \item The communication process was easy
    \item My intentions were hard to transmit
    \item The vocabulary was easy to remember
    \item I felt confused about the robot's dictionary
    \item I was not sure about what to tell to the robot
\end{enumerate}

\noindent\textbf{Speech recognition (FBV only)}
\begin{enumerate}
    \setcounter{enumi}{25}
    \item The robot understood me in my normal volume of speech
    \item I had to shout to be understood
    \item I had to whisper to be understood
    \item The robot understood me in my normal speech velocity
    \item I had to speak too slowly to be understood
    \item I had to speak too fast to be understood
\end{enumerate}

\noindent\textbf{Responsiveness (FBV only)}
\begin{enumerate}
    \setcounter{enumi}{31}
    \item The robot acted fast after I spoke a command
    \item The robot was too slow in acting upon my spoken commands
    \item The robot usually understood me the first time I talked
    \item I had to repeat myself more than once to be understood
\end{enumerate}

\noindent\textbf{Anthropomorphism (FBV only)}
\begin{enumerate}
    \setcounter{enumi}{35}
    \item Talking to the robot felt like talking to a human
    \item Talking to the robot was more difficult than talking to a human
\end{enumerate}

\noindent\textbf{Direction intent (FBV only)}
\begin{enumerate}
    \setcounter{enumi}{37}
    \item I could express my intention of direction easily
    \item The start of the movement was better with direction intent
    \item I felt like the way I expressed intent was complicated
\end{enumerate}

\noindent\textbf{Open-ended questions}
\begin{enumerate}
    \setcounter{enumi}{44}
    \item Additional comments
    \item Did you prefer the interaction with the robot with or without speech recognition? Why?
\end{enumerate}

\section*{ACKNOWLEDGMENT}

The authors would like to thank Bruno Gabriel Cavalcante Lima for laying the foundations of this study.


\bibliographystyle{IEEEtran}
\bibliography{IEEEabrv, OtherAbbrv, bibliography}

\end{document}